\documentclass[pre,aps,showpacs,twocolumn,floatfix]{revtex4}
\usepackage{epsf}
\usepackage[dvips]{color}
\usepackage{graphicx,color,float,amsmath,amssymb}

\begin{document}

\title{Distinguishing Fact from Fiction: Pattern Recognition in Texts\\ Using Complex Networks}
\author{J. T. Stevanak, David M. Larue, and Lincoln~D. Carr\\}
\affiliation{Department of Physics, Colorado School of Mines, Golden, CO 80401, U.S.A.\\}
\date{\today}

\begin{abstract}
We establish concrete mathematical criteria to distinguish between
different kinds of written storytelling, fictional and non-fictional.
Specifically, we constructed a semantic network from both novels and
news stories, with $N$ independent words as vertices or nodes, and
edges or links allotted to words occurring within $m$ places of a
given vertex; we call $m$ the word distance.  We then used measures
from complex network theory to distinguish between news and fiction,
studying the minimal text length needed as well as the optimized word
distance $m$.  The literature samples were found to be most
effectively represented by their corresponding power laws over degree
distribution $P(k)$ and clustering coefficient $C(k)$; we also studied
the mean geodesic distance, and found all our texts were small-world
networks.  We observed a natural break-point at $k=\sqrt{N}$ where the
power law in the degree distribution changed, leading to separate
power law fit for the bulk and the tail of $P(k)$.  Our linear
discriminant analysis yielded a $73.8 \pm 5.15\%$ accuracy for the
correct classification of novels and $69.1 \pm 1.22\%$ for news
stories.  We found an optimal word distance of $m=4$ and a minimum
text length of 100 to 200 words $N$.
\end{abstract}


\maketitle

\section{Introduction}
Both written and spoken languages have complex grammatical structure that dictates how information can be communicated from one individual to another.  Recently, complex network theory has become one of the tools used to study the structure and dynamics of languages~\cite{Sigman, Dorogovtsev, Steyvers, Markosova, Cancho, Chan, Mizraji, Bales, Ferreira, Fuks, Bernhardsson}.  The study of \textit{semantic networks} is an interdisciplinary endeavor spanning many fields, including statistical physics, information theory, linguistics, and cognitive science.  Semantic networks can refer to experimentally testable organic models of functioning in the brain~\cite{vandenberghe1996,andres2004} as well as statistical models of connections and patterns in texts.  Although the two may in fact be related, it is particularly the latter form of semantic network that interests us here; such networks have been constructed from lexicons, dictionaries, prose and poetic fictional literature, and thesaurus networks~\cite{Sigman,Dorogovtsev,roxas2010,Steyvers}.   Dorogovtsev and Mendes have even provided an analytical model for the time-evolution of a language network~\cite{Dorogovtsev}.  Most of these approaches study the topology of a single network, sometimes composed of many samples in order to better understand the structure of language.  However, in the same way that the overall structure and behavior of a language network is important, so too is the use of a language for particular forms of communication important.  Specifically, there are many questions that could be asked about substructures that arise in a language, such as  what is the difference between networks that are composed of samples from different literary styles?  Can we use complex network theory to distinguish between different types of literature?  What is the critical number of words in a piece of literature at which these differences become apparent to the reader, and how might we mathematically model this kind of pattern recognition?  This Article addresses these questions.

{Prior to approaches using semantic networks, there are numerous
  studies doing either the classification of text into a variety of
  subject area, or, closer to the present study, the classification
  into genres.  Text Classification is of particular interest in query
  and retrieval systems, since restricting to texts that are concerned
  with the same topics as the query can improve efficiency and
  correctness.  Typically, some features are extracted from the text,
  such as lists of words or sets of adjacent word stems and their
  frequencies, sometimes with meanings or qualities attached.  Then a
  training set of texts with known classifications is used with some
  methodology such as linear maps to a small set or just one value,
  with cutoff values experimentally determined to give the best
  classification.  Other methodologies include Bayesian Independence
  probabilities, Neural Networks, etc.  Then these classification
  methods are applied to new texts, and their effectiveness measured.
  Useful surveys of the contexts, methods and literature in this area {are}
  \cite{sebastiani2002machine,sebastiani1999tutorial}.  Effectiveness
  varies, as does the complexity of methods and amount of specialized
  knowledge embedded in them.  Because the number of categories
  tends to be somewhat large (the often used Reuters collection having
  nearly 100, and the OHSUMED collection has more than 10,000, with
  multiple categories applicable to each text (see
  \cite{yang1999evaluation}), accuracy rates are not directly
  comparable to a binary genre classification into fact or fiction as
  in the present Article.}

{The narrower field of automated genre
  classification has a now extensive literature as well.  Similar
  considerations apply as for more general text classifications.
  Search and retrieval remain applications of interest, as well as the
  automated organizing of digital libraries.  This is distinct from
  the classification of texts by subject, as texts on the same subject
  may be in different genres.  In \cite{karlgren1994recognizing} we see an
  account of various classifications, including into two categories of
  Informative and Imaginative, which in 500 texts had error rates of
  4\% and 5\% respectively.  Discriminant analysis was run against the
  Brown corpus of English text samples of equal lengths on a score of
  features, such as counts of word lengths, first person pronouns,
  sentences, etc.  The training and test sets were the same, and no
  attempt to estimate estimation variances was made.  When applied to
  a larger number of sub-genres, the error rates went up.  The larger
  number of features adds to the effectiveness of the discriminant
  analysis.  So this is similar in some respects to the present
  Article.  This Article may then be considered as a proof of concept
  that measures derived from complex networks have validity as well as
  those of simple counts or with language or subject specific
  knowledge encoded in the methods.}

{Recall that previous studies explicitly using networks did
  so with a single network.}  This Article departs from those
approaches. In lieu of a single network composed of many literature
samples, we chose to represent each literature sample as its own
undirected, unweighted network.  From these networks, we were able to
compare the power law behavior of degree, clustering, and mean
geodesic distance, all established useful measures in complex network
theory.  Complex network theory has studied many aspects of human
behavior, such as mobility patterns~\cite{gonzalezMC2008}. Here we
study on the basic human activity of storytelling.  We show that
complex network theory can be used to distinguish between storytelling
in different forms: in novels, where the stories are fictional, and in
news stories, where the stories are non-fictional.

Our article is outlined as follows.  In Sec.~\ref{sec:model} we describe the semantic network model that was used to create complex networks from our text samples, providing an explicit example based on a quote from Feynman.  Section~\ref{sec:analysis} details the different kinds of behavior the networks exhibited and our method of exploiting this behavior in order to distinguish between two modes of storytelling.  Section~\ref{sec:conclusions} lists our conclusions as well as outlook for future studies.

\section{Semantic Network Model and Measures}\label{sec:model}

\subsection{Basic Model and Example}

Words and the meanings that they represent are foundational to the grammar of a language, but without a conceptual structure within which to arrange them there can be no communication of coherent thought.  Thus the meaning of a single word is equally important as understanding the context of the sentence, paragraph, extended text, or story in which that word appears.  Although conjugation, the use of singular or plural, and other grammatical forms contribute to the understandability and elegance of a story, nevertheless the essence of the story can be communicated in pidgin English or by a child, without proper use of all grammatical rules. These observations form the qualitative basis for our model.

Using a variation of the model proposed by Cancho and Sole~\cite{Cancho}, we constructed an unweighted, undirected semantic network model that take a single text sample as a network.  Each vertex in the network represents all occurrences of a unique word in the text.  Thus a text of $N_{\mathrm{words}}$ words has in general $N$ vertices, where $N < N_{\mathrm{words}}$ and usually $N \ll N_{\mathrm{words}}$; for example, all appearances of ``and'' in a text would count as a single vertex.  Two vertices are assigned a link or edge between them if the words that they represent appear within $m$ words of each other in the text sample.  We call $m$ the \textit{word distance}.  Since our semantic network is unweighted, once two vertices have an edge, further edges appearing from other occurrences of the same two vertices are not counted.  To avoid focusing on grammatical forms and extract storytelling forms as independent of grammatical usage as possible, we allow a single vertex to not only represent all occurrences of its assigned word, but to also represent all lexical forms of that word.  For example, ``eat'', ``eats'', and ``eaten'' would be represented by the same vertex in our model.  Contractions were considered to be two different words, and all occurrences of `` 's'' were excluded because of the difficulty in distinguishing between the possessive form of an `` 's'' word and the contraction of the word with ``is''.

Of course, more complicated semantic network models are certainly possible.  For instance, one could construct a weighted network.  However, we sought the simplest possible model which could distinguish between fictional and non-fictional written storytelling.

\begin{figure}[htbp]
	\centering
	\includegraphics[width=8.5cm]{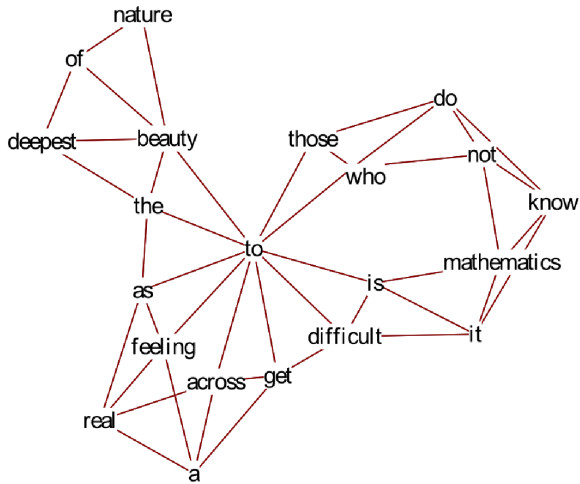}
	\caption{Unweighted undirected semantic network created from a quote by Richard Feynman.  The word distance is taken as $m=2$.\label{toynetwork}}
\end{figure}

In order to illustrate our model we will analyze a quote by Richard Feynman: ``To those who do not know mathematics it is difficult to get across a real feeling as to the beauty, the deepest beauty, of nature...''  If we choose $m=2$ then Fig. \ref{toynetwork} displays the resulting network.  Notice how the word ``beauty'' is assigned to only one vertex.  That vertex is then connected to the $2m$ nearest neighbors of both instances of ``beauty''.  These $2m$ neighbors are ``to'', ``the'', ``the'', and ``deepest'', from the first occurrence and `the'', ``deepest'', ``of'', and ``nature''.  One can see also that there is only one edge between ``the'' and ``beauty'' and no self edges on ``beauty''.  Our model does not allow self loops nor does it allow multiply connected pairs of vertices in order to ensure that the resulting adjacency matrix is indeed unweighted.  Any and all punctuation is also ignored in the text sample.  This results in connections across commas, periods, and other forms of punctuation.  Thus our model considers the overall connectivity of a text independent of individual sentences.

\subsection{Complex Network Measures in Semantic Networks}

The main two network measures that we used were the degree of a vertex and the clustering coefficient of a vertex.  The degree of a vertex is the number of edges associated to the vertex, i.e., its number of connections.  In an unweighted adjacency matrix, the degree ($k$) of the $i$th vertex can be calculated by summing over the $i$th column of the adjacency matrix $a_{ij}$.
\begin{equation}
k_{i}=\sum^{N}_{j=1}{a_{ij}},
\end{equation}
where $N$ is the number of vertices in the network.  The clustering coefficient $C_{i}$ of a vertex describes how well connected the vertex is to the rest of the network by summing over all closed paths involving two other vertices:
\begin{equation}
C_{i}=\sum^{N}_{j,m=1}{\frac{a_{ij}a_{jm}a_{mi}}{k_{i}(k_{i}-1)}}.
\end{equation}
The clustering coefficient is normalized between 0 and 1, with 1 being fully connected and 0 being not connected at all.

In order to extract fundamental features from degree and cluster, we consider the distribution of values over each data set, i.e., each text sample.  For degree we found $P(k)$, the number of vertices having degree $k$, to be the most useful quantity to consider.  Then we fit two separate power laws to $P(k)$ for $k < \sqrt{N}$ and $k > \sqrt{N}$, as we discuss in more detail in Sec.~\ref{sec:analysis}.  The particular distribution chosen and the power law fits were determined after different types of curves had been fit to various distributions.  For example, we also fitted both exponentials and Poisson distributions to $P(k)$, and we considered a single power law fit without the break at $k = \sqrt{N}$; however, all were poor fits for the data set according to their reduced $\chi^2$.  For the clustering we found the distribution $C(k)$ to be most useful.

With our network model, the linguistic significance of the degree distribution $P(k)$ can be interpreted as follows.  The degree distribution can be understood as the fraction of words in the network which appear a certain number of times.  Because a word gains an upper bound of $2m$ connections every time that it appears in the text sample, $k/2m$ is a lower bound on the number of times a word is present in the text, where $k$ is the degree of a vertex.  A word that appears more than once has a finite probability of being next to a word to which it is already connected, thus $\frac{k}{2m}$ may be smaller than the total number of instances of a word.

Different network measures can mean very different things in separate types of networks.  For semantic networks, the clustering coefficient of a word can be interpreted as a measure of the breadth of the word's usage or the number of times it occurs within a set phrase.  If a word occurs frequently within a certain phrase, then it will tend to have a higher clustering coefficient than a word such as ``a'', ``an'', or ``the'', which are all used very often without regard to any specific phrase.  The higher clustering coefficient value arises from the fact that a word will have more highly connected neighbors due to the multiple instances of the entire phrase in the text sample.  This can be highly affected by the word distance $m$ of the network.  Phrase neighbors will not be as well connected to each other in an $m=1$ network as they would be in an $m=2$ or higher network.  This could also be described in terms of a hierarchy.  Phrases represent modular structures in the network which have higher clustering than the network average~\cite{Ravasz2003}.  Articles such as ``a'', ``an'', and ``the'' will tend to not be related to any specific hierarchy of terms, but may be found in many.

Another factor that could affect a word's clustering coefficient is the range of terms that can commonly be associated with the word.  For instance, the word ``biological'' can be applied to a narrower range of terms than the word ``blue''.  A house, car, bird, pencil, etc. can be blue, while fewer of those could be considered to be biological.  This specificity inherent in the meaning of a term as it applies to a field or group of other terms can increase the clustering coefficient of that word.  If a term has a narrower range of meaning or a smaller subset of words which it can be related to then it will have a higher clustering coefficient, whereas words that can be used more interchangeably will tend to have lower clustering coefficients.  This relationship is encapsulated in the power-law $C(k)$, the degree dependence of the clustering coefficient.  The range of meaning of a word translates into the word's ability to appear within multiple word hierarchies in the network.  In the previous example, ``biological'' can relate to a smaller hierarchy, (given that specific set of control terms) than the word ``blue''.  Again, this tendency can be affected by the chosen word distance.  If the parameter $m$ grows, the clustering coefficient of a vertex will be less affected by the previously mentioned literary nuances, and more affected by the number of times that it appears in the sample.

As an example of another measure we considered, consider the mean geodesic path of the network, also called the closeness, defined as the average shortest path between any two vertices.  The mean geodesic distance $l$ of the network is~\cite{Newman2003}:
\begin{equation}
l=\frac{1}{\frac{1}{2}N(N+1)}\sum_{i\geq j}{d_{ij}},
\label{eqn:geodesic}
\end{equation}
where $d_{ij}$ is the geodesic distance between two vertices $i$ and $j$.  When $l$ is on the order of $\log_{10} N$ then the network is said to have the small world effect~\cite{Newman2003}.  We will discuss $l$ as an alternate measure in Sec.~\ref{sec:analysis}.

\section{Data Analysis of Fictional and Non-Fictional Storytelling}\label{sec:analysis}

To explore storytelling we chose two particular fictional and non-fictional literary forms, novels and news stories.  We contrast our data sets to more obviously distinguishable literary forms such as poetry and prose~\cite{roxas2010}, which have significant differences in structure and often in grammar as well.  Our initial analysis was not performed on the entire length of each sample, but was instead limited to between 500-1000 words.  Subsequently we more finely controlled the size of the networks to determine the number of words required for our analysis to distinguish between the different storytelling modes.

\subsection{Fitting Methods and Fisher's Linear Discriminant}
\begin{figure}[htbp]
	\centering
	\includegraphics[width=8.5cm]{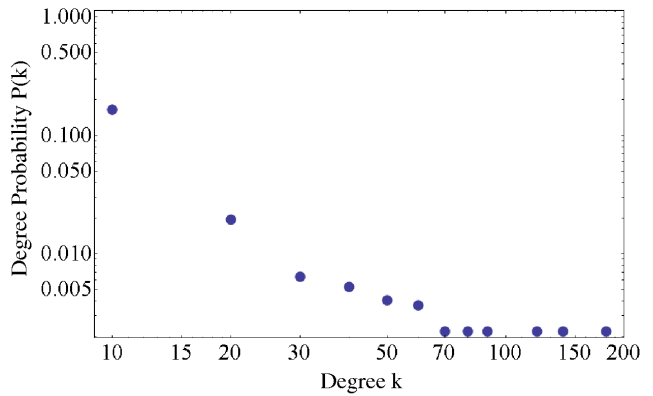}
	\caption{Log-log plot of the degree distribution of a news story.  A running average is used to bin every 10 points in $k$.  Note the break at $k\simeq \sqrt{N}\simeq 65$ necessitating separate power law fits in regions $k<\sqrt{N}$ and $k>\sqrt{N}$.\label{degree}}
\end{figure}

The word networks that we constructed displayed power-laws in their degree distributions, as discussed in Sec.~\ref{sec:model}, of form
\begin{equation}
P(k)= A_\sigma k^{-\gamma_\sigma},
\label{eqn:degreePowerLaw}
\end{equation}
where $P(k)$ is the number of occurrences of degree $k$ within a single text sample, $A_\sigma$ is a proportionality constant, $\gamma_\sigma$ is the power-law coefficient~\cite{Newman2003, Boccaletti, Ravasz2002, Bianconi}, and $\sigma\in\{1,2\}$ refers to two distinct regions.  We found that the semantic networks exhibited two distinct power-law regions, as can be seen in Fig.~\ref{degree}. The division between the two typically occurred at $k=\sqrt{N}$, where $N$ is the total number of words in the network.
This result is similar to that of Cancho and Sole, who also found two
power-law regimes in their network composed of the British National
Corpus.  However, our large $k$ region was always found to have a
weaker power law than our small $k$ region, $\gamma_2 < \gamma_1$,
whereas Cancho and Sole found the opposite.  This is due to the
difference in the size of the networks being used.  Where their Corpus
network had a maximum degree on the order of $10^{5}$, ours had degree
on the order of $10^{2}$, which is due to the restricted size of our
analysis.  {Given the size of the texts that were analyzed,
    degrees $k$ above about $\sqrt{N}$ had dropped to counts that were
    nearly constantly 1.  Even though it is plausible by Zipf's law
    that for sufficiently large samples a single power law would hold
    farther past this point, the integral nature of the degree
    frequencies makes a smooth fit between these halves problematic.  In fact, a natural break at $k\sim\sqrt{N}$ is typical of small world networks: high degree words for $k > \sqrt{N}$ represent common words which form the common framework of the English language, while low degree words below $\sqrt{N}$ are rare words.  We expect that common words like ``and,'' ``the'', etc. are used quite differently than specialized words like ``biological.''}
  The clustering coefficient distribution was found via various fits
  to be best represented exhibit by a single power law of form
\begin{equation}
C(k)=A_3 k^{-\gamma_3},
\label{eqn:clusteringPowerLaw}
\end{equation}
where $C(k)$ represents the set of all vertices $i$ with clustering coefficient $C_i$ and degree $k_i$.  In Eqs.~\ref{eqn:degreePowerLaw}-\ref{eqn:clusteringPowerLaw} one can normalize $A_1,A_2,A_3$; however, as we will only utilize the exponents normalization is irrelevant for our analysis.
\begin{figure}[htbp]
	\centering
	\includegraphics[width=8.5cm]{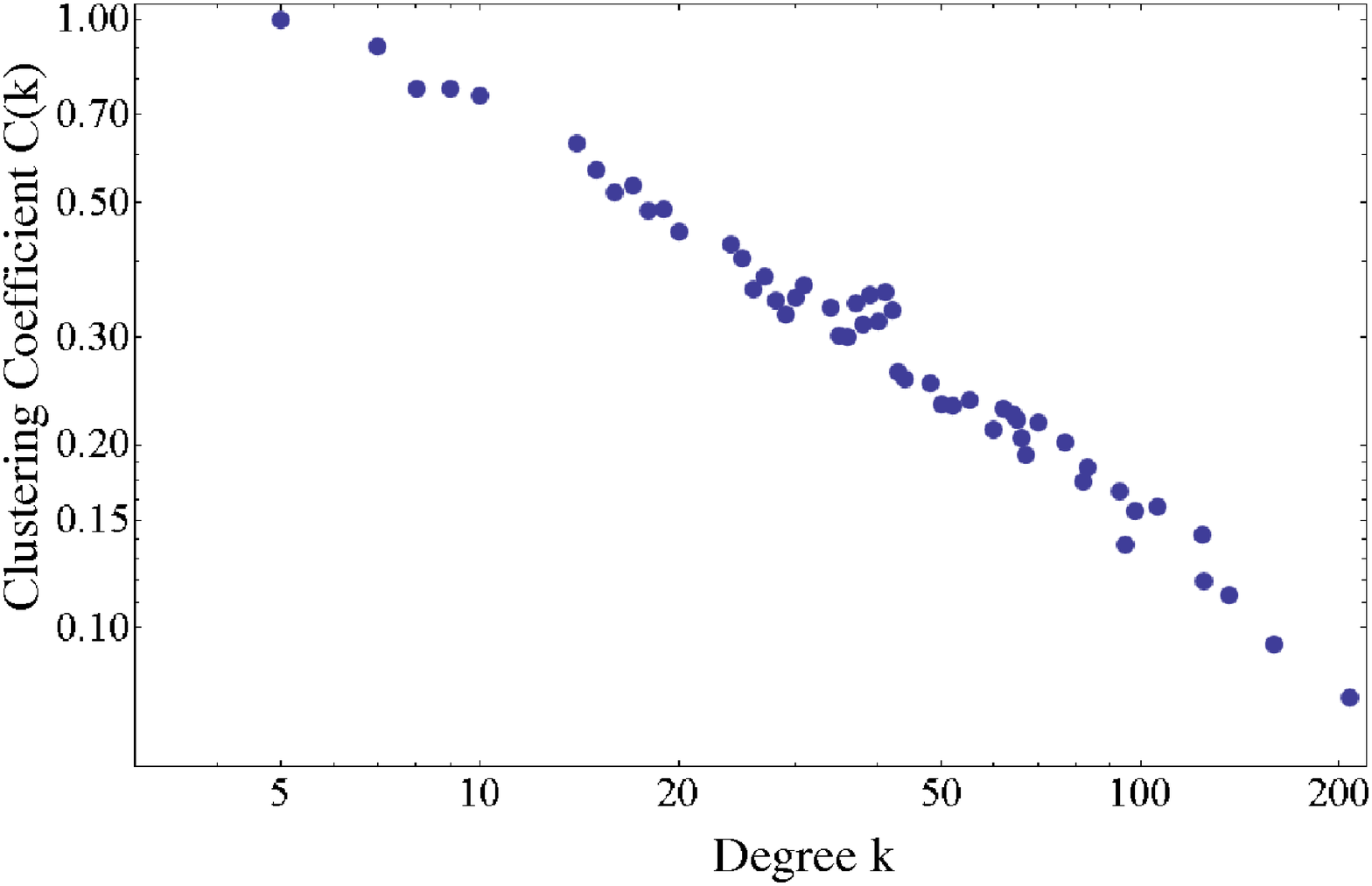}
	\caption{Log-log plot of the degree dependence of the clustering coefficients of a novel. Unlike in the case of degree distribution in Fig.~\ref{degree}, here there is only a single power law evident.\label{cluster}}
\end{figure}

\begin{figure}[htbp]
	\centering
	\includegraphics[width=8.5cm]{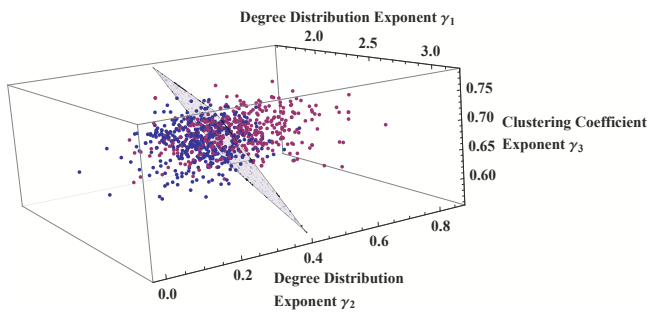}
	\caption{(color online) Novels (dark blue points) and news
          stories (lighter purple points) represented by three
          power-law exponents, $\gamma_{1}$ and $\gamma_{2}$ from the
          degree distribution, and $\gamma_{3}$ from the clustering
          coefficient\label{exponentspace}{, together
            with the best planar separation}.}
\end{figure}

There are many parameters and values that can be chosen to represent
and analyze networks~\cite{Fortunato, Sokolov, Chan, Costa, Masucci,
  Ma, Santiago, Mizraji, Costa2007, Zhang, Caldeira, Bales, Bianchini,
  Szabo, Benito}.  For our simplest possible semantic network we
sought the minimal set of measures needed to distinguish between
fictional and non-fictional written storytelling, and found the power
law exponents $\gamma_{1},\gamma_{2},\gamma_{3}$ to be optimal.
{Measures that
  were tried and discarded in attempting to find a minimal set
  included magnitudes of the power law fits, small word lengths,
  average sentence lengths and average distance between verbs.}
{Additional
  parameters in the Linear Discriminant Analysis would have
  strengthened the accuracy at the expense of simplicity.  Further
  more detailed analysis is warranted in the future.}
{But a Principal Component Analysis
  would not take advantage of the known classifications, and hence may
  give misleading conclusions as to the importance of different
  parameters.}  Figure~\ref{exponentspace} shows a distribution over
all data sets, with each point representing all three exponents for a
novel or a news article.  {The plane
  represents the division that Fisher's Linear Discriminant, discussed
  later in this section, induces on the data set.}

{Because small values of $m$, such as the $m=4$ that
  was found to be optimal, induces oscillations in the degree
  probability as a function of the degree $k$ out to $m$ away from
  the $k$s that are divisible by $2m$, binning the degrees on the order
  of every $2m$ removes this signal that is an artifact of how the
  network is constructed rather than of distinctions in types of
  language being employed, and is therefore reasonable.  Using a
  cumulative distribution instead of the one used here would not
  remove this oscillation.  These oscillations could be more closely
  analyzed and subtracted out in the future, which would remove the
  need to bin.}  Before fitting a curve, the degree distributions were
binned so as to smooth them.  The data values in certain ranges were
averaged.  This technique allows us to look at different scales of
data sets.  The bin width was set equal to $2m$; this bins the
distributions so that the probabilities represented were the
probabilities of selecting a vertex out of the network whose assigned
word appears at least $\frac{k}{2m}$ times in the text sample.  This
representation of the degree distributions is useful because the
values of $P(k)$ for $k$ that is not an integer multiple of $2m$ are
lower as compared to those for integer multiples of $2m$.  The
discrepancy between these values arises because non-integer multiples
of $k$ correspond to words that appear next to the same word multiple
times in the text sample.  For example, in an $m=2$ network, the
probability of having a vertex with a degree of 5 is lower than the
value expected by the power-law because in order to have a degree of
5, a word must appear twice in the sample and, in its second
appearance, be next to three of its four previous neighbors.  This
event is one of lower probability when compared with the probability
of appearing next to four completely new words in a second appearance.

Once the characteristic values of $\gamma_1,\gamma_2,\gamma_3$ were calculated, we needed a method to distinguish between the two groups of data.  At first, we averaged the data points together to produce the center of each distribution.  A new data point could then be categorized as a novel or news article depending upon which center was geometrically closest to the new data point.  The approach could then be tested by taking data points of known category and running the algorithm to see if the data was placed correctly.  This approach achieved results no better than blindly classifying the text samples, a 50-50 accuracy rate, as can be guessed from a careful consideration of Fig.~\ref{exponentspace}.

\begin{figure*}[!t]
	\centering
	\includegraphics[width=\textwidth]{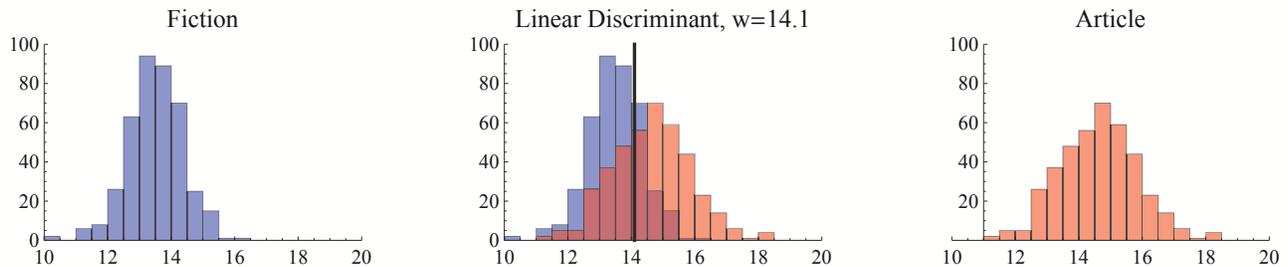}
	\caption{{Distribution of the univariate value from Fisher's Linear Discriminant analysis for novels and news stories, showing the separation between them and a useful discrimination (black line in center panel).}\label{FischerLDA}}
\end{figure*}
When this method proved ineffective, we turned to Fisher's Linear Discriminant in order to distinguish between the two groups~\cite{Johnson}.  The linear discriminant transforms a multivariate observation of a group into a univariate one in order to be able to classify data points into different groups.  The single value $y_{0}$ representing a multi-dimensional data point $\vec{x}_{0}$ can be found by using
\begin{equation}
y_{0}=(\overline{x}_{1}-\overline{x}_{2})^{T}S_{\mathrm{pooled}}^{-1}\vec{x}_{0},
\end{equation}
where $\overline{x}_{1}$ and $\overline{x}_{2}$ are the column vectors representing the average of data groups 1 and 2, respectively, and $S_{\mathrm{pooled}}^{-1}$ is the inverse of $S_{\mathrm{pooled}}$, which is a matrix defined as
\begin{eqnarray}
S_{\mathrm{pooled}} & = & \frac{1}{(n_{1}+n_{2}-2)} \nonumber \\
&& \times \Big\{ \sum_{j=1}^{n_{1}}(\vec{x}_{1j}-\overline{x}_{1})[(\vec{x}_{1j}-\overline{x}_{1})]^{T} \nonumber \\
&& + \sum_{j=1}^{n_{2}} (\vec{x}_{2j}-\overline{x}_{2}) [(\vec{x}_{2j}-\overline{x}_{2})]^{T} \Big\},
\end{eqnarray}
where $\vec{x}_{1j}$ and $\vec{x}_{2j}$ are the individual column vectors representing individual data points in groups 1 and 2, respectively, and $n_{1}$ and $n_{2}$ are the number of data points in their respective groups.  In order to distinguish between two groups of distinct data, one must first define the univariate midpoint $w$ between the two.
\begin{equation}
w=\frac{1}{2}(\overline{x}_{1}-\overline{x}_{2})^{T}S_{\mathrm{pooled}}^{-1}(\overline{x}_{1}+\overline{x}_{2}),
\end{equation}
To perform the analysis, two sets of control data are required in order to define the two groups.  When these groups have been selected, one can then calculate the univariate value of a data point not in either the control group.  If $y_{0}\geq w$ then the data point belongs in group 1, otherwise it belongs in group 2.

{In Figure~\ref{exponentspace}, the
  scatterplot has been rotated to best exhibit the separation between
  the two datasets, and the plane
  $2.3\gamma'_1+8.2\gamma'_2+2.2\gamma'_3=14.1$ (where
  $\gamma'_i=\gamma_i/\Delta\gamma_i$ has been normalized by the
  dispersion of each variable) has been included that separates the
  points based on Fisher's Linear Discriminant analysis.  In
  Figure~\ref{FischerLDA}, the distribution of the univariate values for each of
  the two groups of data is given separately, and combined in the
  center, together with the dividing value of $w=14.1$.  We can see
  that these two distributions are mounded, with different means and
  variances, and are sufficiently distinct that this dividing point
  has predictive value.}

\subsection{Novels vs. News Stories}

\begin{figure}[htbp]
	\centering
	\includegraphics[width=8.5cm]{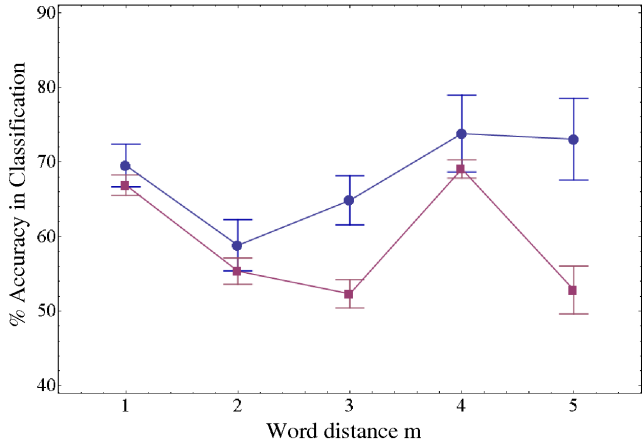}
	\caption{(color online) Percent accuracy of the linear discriminant analysis in distinguishing between fictional (blue, upper curve) and non-fictional (purple, lower curve) storytelling, as a function of the word distance $m$ used in the semantic network model. The curves are a guide to the eye; error bars are obtained from the bootstrapping method.\label{accuracy}}
\end{figure}

We gathered a total of 400 random novels from www.gutenberg.org and
400 random news stories from www.npr.org.  In each category, 200
randomly selected samples were used as a control data set, and
Fisher's Linear Discriminant was used on the other 200 samples.  For
the news stories we focused on reporting on current events as the best
example of non-fictional storytelling; for novels we took only modern
writing, i.e., 20th century, so as to remove effects of archaic
language.  An accuracy of classification for each category was
calculated based upon the number of correct classifications made by
the discriminant analysis divided by the total number of text samples
in that category.  Figure~\ref{accuracy} displays the accuracy of the
linear discriminant as a function of word distance $m$.  The greatest
accuracy that we calculated was $73.8 \pm 5.15\%$ for the correct
classification of novels and $69.1 \pm 1.22\%$ for the classification
of news stories.  This accuracy was achieved at $m=4$ in our semantic
network model.  {These values are comparable to those found by
  \cite{roxas2010} in their disambiguation of poetry and prose by
  similar methods.}  We take these values as the peak accuracies not
only because the novels have the largest accuracy at this
connectivity, but also because the relative distance between the
accuracies of the novels and the news stories is smallest.  Not only
one, but both of the accuracies must be larger than approximately 60\%
in order for that particular value of $m$ to be considered a better
method than just randomly choosing a category for an unknown
literature sample.  At all stages of our analysis proper methods for a
blind study were implemented in our code.

These results came from an analysis that only took into consideration the two exponents $\gamma_1,\gamma_2$ from the degree distribution and the exponent $\gamma_3$ from the clustering coefficient distribution.  We considered a number of other possibilities, of which we describe one here as follows.  We ran an analysis that used the mean geodesic path (or small world length) $l$ in addition to the three exponents.  This analysis yielded results that agreed to within 0.01\% of the original analysis.  The reason for this is that all of the mean geodesic paths were very similar.  We found that all texts were small world networks, meaning that the mean geodesic path of each of the networks was $l \approx \log_{10} N$.  Since $N$ was similar in magnitude for all of the text samples the mean geodesic path for each of them was similar as well.

The $m=1$ semantic networks displayed the second highest accuracy.  This can be interpreted to mean that the majority of the information implied by the presence of a word is contained within the connection to that word's nearest neighbors, which is indeed true.  After $m=4$, the accuracy begins to drop, as one can observe in the $m=5$ study in Fig.~\ref{accuracy}.  As the word distance of the network increases, the number of false connections also increases.  That is to say, as the connectivity grows, the number of meaningless or incorrect connections occurs.  Though there are many words in a sample that could be considered to be connected to words relatively far away, such as words in a distantly placed adverbial or adjectival clause, most words can't be considered to have such connections.  Moreover, the more connected a network becomes the closer it gets to being a maximally or fully connected network, whose topology does not consist of power-laws, and cannot be significantly distinguished from another fully connected network.

\begin{figure}[!t]
	\centering
    \vspace*{6cm}\includegraphics[angle=-90,width=0.48\textwidth]{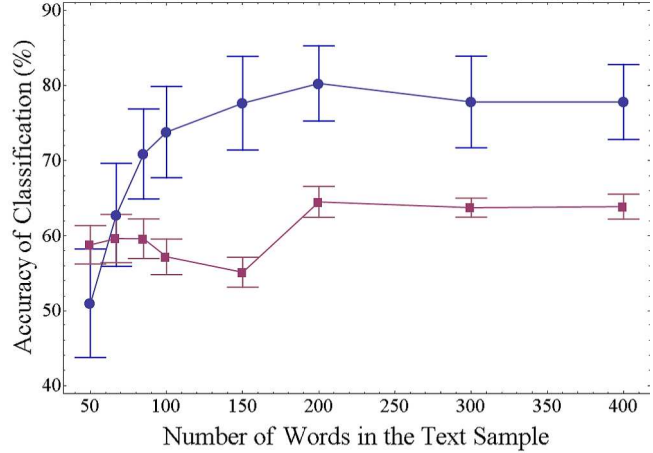}
	\caption{Accuracy of categorical classification of texts versus the sample size of the network.  The word distance was held constant at $m=4$.  Maximal accuracy occurs at about $N_{\mathrm{words}}=$ 200 words, although categories can still be distinguished even at just 75 words.  Note that in general the number of independent words or vertices $N$ in our semantic network model in general obeys $N \ll N_{\mathrm{words}}$.  The color scheme and method is the same as that of Fig.~\ref{accuracy}.\label{lengths}}

\end{figure}

In addition to testing the accuracy of different word distances, we also checked how accurate the analysis became when the text samples were shortened.  The human mind can quickly determine fictional vs. non-fictional story-telling as represented by novels and news stories, typically within a few lines.  We tested our semantic network model using different lengths of text to see how well it distinguished between the two control groups.  The model was tested with text sample lengths of 50 to 400 words, with the connectivity held fixed at $m=4$.  Fig.~\ref{lengths} summarizes the results.  As the plot shows, the accuracies of classification begin near 50$\%$ and then improve starting at 75 words.  At 200 words, the networks have approximately reached their highest accuracy.  If this is compared with the comprehension of an adult, the result is reasonable. By the time a person has read about one paragraph of a text, the fictional aspect of a story is clear.  We clarify that the number of vertices $N$ is usually much less than the number of words $N_{\mathrm{words}}$.

{ By the relative sizes of
  the $\gamma'_i$ coefficients in the equation of the plane of
  separation, we see that the largest contributor to accurate
  distinguishing of these two genres comes from $\gamma_2$, the degree
  distribution power law exponent for the larger regime of degrees.
  Interpreting this observation calls for additional study, as these
  larger degree values are infrequent, often occurring only once.  But
  one can roughly say that news articles generally have a higher value
  for $\gamma_2$, and this is associated with a steeper drop-off in
  frequencies of the degrees of the nodes for the larger degrees.
  This may come about if articles have a larger number of high degree
  words, which is to say that they have more repeated or common words.
  This rings true as fiction writers consciously try to vary their
  vocabulary, while writers of news articles have collections of stock
  words and phrases.  This hypothesis could be tested by independent
  analysis of these texts and their word frequencies.  However one
  attempt to distinguish these genres by word frequencies is detailed
  below with negligible accuracy.}

In order to achieve an estimate of the error inherent to our analysis methods for our semantic network, a bootstrapping method was used on the discriminant analysis.  This approach is an iterative method used to deduce the size of a distribution of a random variable $x_{1}$ strictly from numerical data without any knowledge of the theoretical distribution of $x_{1}$~\cite{Efron}.  The bootstrap method takes a random sample of size $n$ of the total number of observations of $x_{1}$, performs the desired analysis, takes another random sample from the observations of $x_{1}$, and repeats the entire process until it has a set of results from the desired calculation from which one can calculate the mean and standard deviation.  We used twice the standard deviation as the error for each of our data sets and used the mean value as the reported value.  With our discriminant analysis, the bootstrapping algorithm randomly chose 200 of the 400 text samples of each group (news stories and novels) and calculated the accuracy of classification for each.  We chose to use 200 random samplings in our error analysis. This provided us with a list of accuracies from which we could calculate the mean and the standard deviation.  The mean and standard deviations of the bootstrap method converged with increasing iterations.  All error bars in figures are obtained via the bootstrap approach.

{To compare this method with that of a
  simple statistic that does not make use of the complex network
  theory measures, we analyze, for example, the word frequencies in
  articles and fiction.  After breaking the chosen text segment into
  words, we find the probability of each, and rank them from most
  frequent down.  An established empirical result, colloquially known
  as Zipf's law, has that this distribution is well approximated by a
  power law \cite{zipf1935psycho}.  If we fit a power-law to this data
  (and the fit is visually and by norm measure reasonable), and use
  the same discrimination technique as above together with the
  bootstrapping algorithm, we obtain an accuracy of $48.8 \pm 7.4\%$
  for the correct classification of novels and $58.4 \pm 7.1\%$ for
  the classification of news stories.  The worse-than-chance showing
  for fiction is a consequence of the empirical observation from this
  sample that the distribution of the fitted exponents from the
  power-law are approximately symmetrical for articles while they are
  positively skewed for fiction.  Using the average of the medians
  rather than the means in the discrimination improves the accuracy
  for fiction to slightly over chance, while decreasing the accuracy
  for articles to closer to chance.  Other choices of a discrimination
  value could be made to different effects.  But this particular
  simple statistic is noticeably inferior in distinguishing fiction and
  non-fiction as compared with the complex network based measure
  highlighted in this Article.}

\section{Conclusions}\label{sec:conclusions}

We have shown that a combination of established complex network theory measures and multivariate statistical analysis can be used to distinguish between different types of storytelling, fictional and non-fictional.  Specifically in our analysis, an unweighted, undirected network model was used to distinguish between novels and news stories.  Fisher's linear discriminant was used to classify random samples of known novels and news stories.  In this way we were able to calculate an accuracy of classification for our model using the linear discriminant and calculate error bars based on an iterative bootstrap algorithm.

We considered the simplest possible semantic network model, an undirected unweighted network which assigns vertices to independent words independent of their conjugation, pluralization, or other grammatical structure, and assigns edges whenever two words are within $m$ places of each other.  The ensuing networks constructed from novels and news stories have collectively different values for the exponents in the power-laws of both their degree distributions and their clustering coefficients.  These exponents can be used to classify the two literature types.  With the use of the linear discriminant, one can test to see if a story of unknown literature type is fiction (a twentieth century novel) or non-fiction (a news story on current events).  This same method can also be applied to text samples of known category in order to calculate the accuracy with which the model can distinguish between the two types of storytelling.  The highest value for the accuracy of classification for the two literature types was found at a word distance of $m=4$, giving accuracies of $73.8\pm5.15\%$ and $69.1\pm1.22\%$ for the novels and news stories respectively.  Categories could be distinguished beginning at about 75 words and were maximally accurate at about 200 words.

A less coarse semantic network model might lead to substantially improved accuracy in future studies.  A weighted or directed network approach may be able to provide more meaningful structure in the placement or relationships of words.  Other tools could also be used in the analysis such as community detection algorithms in order to further enhance the boundaries between novels and the news stories.

We thank William C. Navidi for discussions of statistical analysis and Norman Eisley for helping us recognize that storytelling is as fundamental a human behavior as, for example, social networks or mobility.


\begin{thebibliography}{10}

\bibitem{Sigman}
M. Sigman and G. Cecchi, Proc. Natl. Acad. Sci. USA {\bf 99},  17421747
  (2002).

\bibitem{Dorogovtsev}
S. Dorogovtsev and J. Mendes, Proc. R. Soc. Lond. B {\bf 265},  2603  (2001).

\bibitem{Steyvers}
M. Steyvers and J. Tenenbaum, Cognitive Science {\bf 29},  41  (2005).

\bibitem{Markosova}
M. Markosova, Physica A {\bf 387},  661  (2008).

\bibitem{Cancho}
R.~F. Cancho and R. Sole, Proc. R. Soc. Lond. B {\bf 268},  2261  (2001).

\bibitem{Chan}
K.~Y. Chan and M.~S. Vitevitch, Journal of Experimental Psychology-Human
  Perception and Performance {\bf 35},  1934  (2009).

\bibitem{Mizraji}
E. Mizraji and J.~C. Valle-Lisboa, Medical Hypotheses {\bf 68},  347  (2008).

\bibitem{Bales}
M.~E. Bales and S.~B. Johnson, Journal of Biomedical Informatics {\bf 39},  451
   (2006).

\bibitem{Ferreira}
A.~A.~A. Ferreira, G. Corso, G. Piuvezam, and M.~S. C.~F. Alves, Physica A {\bf
  387},  2365  (2008).

\bibitem{Fuks}
H. Fuks and M. Krzeminski, J. Phys. A: Math. Gen. {\bf 42},  375101  (2009).

\bibitem{Bernhardsson}
S. Bernhardsson, L.~E.~C. da~Rocha, and P. Minnhagen, Physica A {\bf 389},  330
   (2010).

\bibitem{vandenberghe1996}
R. Vandenberghe {\it et~al.}, Nature {\bf 383},  254  (1996).

\bibitem{andres2004}
A. Pomi and E. Mizraji, Phys. Rev. E {\bf 70},  066136  (2004).

\bibitem{roxas2010}
R.~M. Roxas and G. Tapang, Int. J. Mod. Phys. C {\bf 21},  503  (2010).

\bibitem{gonzalezMC2008}
M.~C. Gonzalez, C.~A. Hidalgo, and A.-L. Barabasi, Nature {\bf 453},  779
  (2008).

\bibitem{Ravasz2003}
E. Ravasz and A.~L. Barabasi, Physical Review E {\bf 67},  026112  (2003).

\bibitem{Newman2003}
M.~E.~J. Newman, Siam Review {\bf 45},  167  (2003).

\bibitem{Boccaletti}
S. Boccaletti {\it et~al.}, Physics Reports {\bf 424},  175  (2006).

\bibitem{Ravasz2002}
E. Ravasz and A.~L. Barabasi, Reviews of Modern Physics {\bf 74},  47  (2002).

\bibitem{Bianconi}
G. Bianconi and A.-L. Barabasi, Phys. Rev. Lett. {\bf 86},  5632  (2001).

\bibitem{Fortunato}
S. Fortunato, Physics Reports {\bf 486},  75  (2004).

\bibitem{Sokolov}
I.~M. Sokolov and I.~I. Eliazar, Physical Review E {\bf 81},  026107  (2009).

\bibitem{Costa}
L. Antiqueira, O.~N. Oliveira, L. Costa, and M.~D.~V. Nunes, Information
  Sciences {\bf 179},  584  (2009).

\bibitem{Masucci}
A.~P. Masucci and G.~J. Rodgers, Advances in Complex Systems {\bf 12},  113
  (2008).

\bibitem{Ma}
H.~L. Y.~Ma and X. Zhang, Physica A {\bf 388},  4669  (2009).

\bibitem{Santiago}
A. Santiago and R.~M. Benito, Physica A {\bf 387},  2365  (2008).

\bibitem{Costa2007}
L. Antiqueira, M.~G.~V. Nunes, O.~N. Oliveira, and L. Costa, Physica A {\bf
  373},  811  (2007).

\bibitem{Zhang}
H. Zhang and M.~K.~M. Rahman, Expert Systems with Applications {\bf 36},  12023
   (2009).

\bibitem{Caldeira}
S.~M.~G. Caldeira {\it et~al.}, European Physical Journal B: Condensed Matter
  and Complex Systems {\bf 49},  523  (2006).

\bibitem{Bianchini}
D. Bianchini {\it et~al.}, Journal of Biomedical Informatics {\bf 39},  451
  (2006).

\bibitem{Szabo}
G. Szabo, M. Alava, and J. Kertesz, Physical Review E {\bf 67},  056102
  (2003).

\bibitem{Benito}
A. Santiago and R.~M. Benito, Physica A {\bf 14},  2941  (2009).

\bibitem{Johnson}
R.~A. Johnson and D.~W. Wichern, {\em Applied Multivariate Statisitical
  Analysis}, 6 ed. (Prentice Hall, New Jersey, 2007).

\bibitem{Efron}
B. Efron and R.~J. Tibshirani, {\em An Introduction to the Bootstrap} (Chapman
  \& Hall/CRC, Boca Ration, FL, 2006).

\bibitem{sebastiani2002machine}
F.~Sebastiani, ACM Comput. Surv. {\bf 34}, 1 (2002).

\bibitem{sebastiani1999tutorial}
F.~Sebastiani, Proceedings of ASAI-99 {\bf 1}, 7 (1999).

\bibitem{yang1999evaluation}
Y.~Yang, Information retrieval {\bf 1}, 69 (1999).

\bibitem{karlgren1994recognizing}
J.~Karlgren and D.~Cutting, Proceedings of the 15th Conference on Computational Linguistics {\bf 21} 1071 (1994).

\bibitem{zipf1935psycho}
G.~K.~Zipf, {\em The Psycho-Biology of Language: An Introduction to Dynamic Philology} (Houghton Mifflin, New York, 1935).

\end{thebibliography}

\end{document}